\pdfoutput=1

\documentclass[11pt]{article}
\usepackage[numbers]{natbib}
\usepackage[preprint]{acl}

\usepackage{times}
\usepackage{latexsym}
\usepackage{placeins}
\usepackage{enumitem}
\usepackage{float}
\usepackage{xcolor}
\definecolor{green}{RGB}{198,239,206}

\usepackage[T1]{fontenc}

\usepackage[utf8]{inputenc}

\usepackage{microtype}

\usepackage{inconsolata}

\usepackage{graphicx}

\usepackage{caption}
\usepackage{listings}
\usepackage{multirow}

\usepackage{booktabs}
\usepackage{multirow}
\usepackage{graphicx}
\usepackage{colortbl}
\usepackage{xcolor}

\usepackage{amssymb}

\title{Auto-Eval Judge: Towards a General Agentic Framework for Task Completion Evaluation}

\author{
 \textbf{Roshita Bhonsle\textsuperscript{1*}},
 \textbf{Rishav Dutta\textsuperscript{1*}},
 \textbf{Sneha Vavilapalli\textsuperscript{1*}},
 \textbf{Harsh Seth\textsuperscript{1*}},
\\
 \textbf{Abubakarr Jaye\textsuperscript{2*}},
 Yapei Chang\textsuperscript{1, 3},
 Mukund Rungta\textsuperscript{2},
 Emmanuel Aboah Boateng \textsuperscript{2},
\\
 Sadid Hasan\textsuperscript{2},
 Ehi Nosakhare\textsuperscript{2},
 Soundar Srinivasan\textsuperscript{2}
\\
\\
 \textsuperscript{1}University of Massachusetts Amherst,
 \textsuperscript{2}Microsoft Corporation,
 \textsuperscript{3}University of Maryland
\\
}

\begin{document}
\maketitle

\begin{abstract}
The increasing adoption of foundation models as agents across diverse domains necessitates a robust evaluation framework. Current methods, such as LLM-as-a-Judge, focus only on final outputs, overlooking the step-by-step reasoning that drives agentic decision-making. Meanwhile, existing Agent-as-a-Judge systems, where one agent evaluates another’s task completion, are typically designed for narrow, domain-specific settings. To address this gap, we propose a generalizable, modular framework for evaluating agent task completion independent of the task domain. The framework emulates human-like evaluation by decomposing tasks into sub-tasks and validating each step using available information, such as the agent’s output and reasoning. Each module contributes to a specific aspect of the evaluation process, and their outputs are aggregated to produce a final verdict on task completion. We validate our framework by evaluating the Magentic-One Actor Agent on two benchmarks, GAIA and BigCodeBench. Our Judge Agent predicts task success with closer agreement to human evaluations, achieving 4.76\% and 10.52\% higher alignment accuracy, respectively, compared to the GPT-4o based LLM-as-a-Judge baseline. This demonstrates the potential of our proposed general-purpose evaluation framework.
\def\thefootnote{*}\footnotetext{Equal Contribution}\def\thefootnote{\arabic{footnote}}
\footnote{Code will be made available on GitHub.}
\end{abstract}

\begin{figure}[ht]
  \centering
   \includegraphics[width=1.0\linewidth]{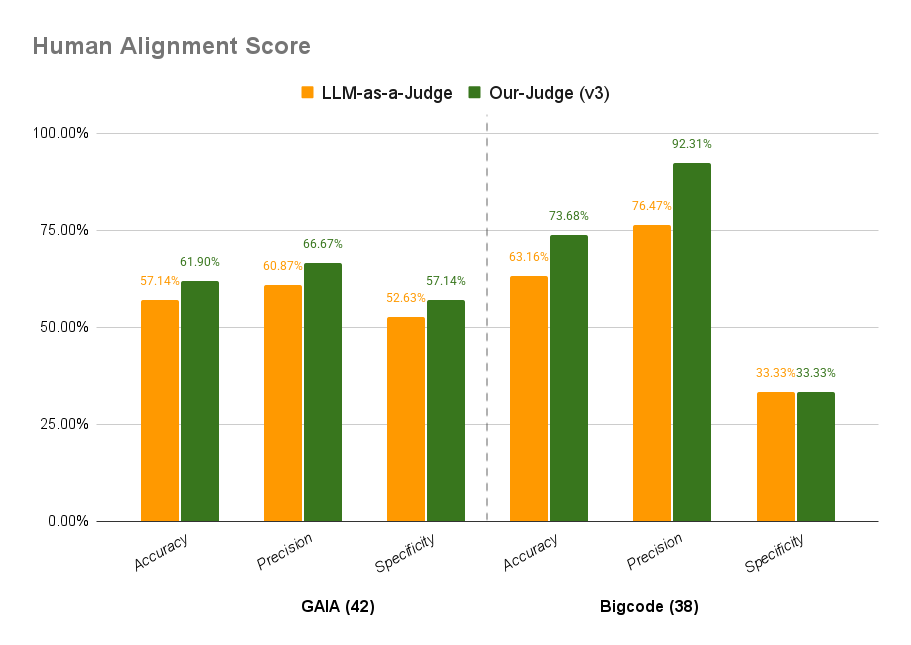}
   \caption{Performance Comparison}
   \label{fig:v3-preformance}
\end{figure}

\section{Introduction}

\paragraph{Problem and Motivation}
We propose a general-purpose, scalable, and modular Agent-as-a-Judge evaluation framework designed to assess agentic task performance with minimal human oversight, while maintaining robustness across diverse domains and datasets.

Human evaluation continues to serve as the de facto standard for determining task success in agentic systems. However, it is inherently limited by its high cost, subjectivity, and lack of scalability posing a critical obstacle to widespread deployment of autonomous agents in real-world settings.

Recent advances have explored leveraging large language models (LLMs) and autonomous agents as evaluators. Notable examples include the LLM-as-a-Judge paradigm \citep{zheng2023judgingllmasajudgemtbenchchatbot} and Meta's Agent-as-a-Judge framework \citep{zhuge2024agentasajudgeevaluateagentsagents}. While promising, these approaches exhibit key limitations: (1) they operate primarily on final outputs and ignore intermediate reasoning traces; (2) they often rely on post hoc human adjudication for ground-truth validation; and (3) they are narrowly scoped, often bound to specific task domains or evaluation protocols.

The following definitions will be used consistently throughout this paper:
\begin{itemize} [itemsep=-2pt, topsep=1pt, leftmargin=12pt]
\item \textbf{$Actor$}: the agent responsible for executing a task through sequential reasoning and interaction.
\item \textbf{$Judge$}: the agent tasked with evaluating the $Actor$’s performance by analyzing both its intermediate reasoning states and final output artifacts.
\end{itemize}

To address these shortcomings, we formalize the evaluation setup by introducing two core agentic roles and framing our contributions around two central research questions:

\paragraph{Research Questions}
\begin{itemize} [itemsep=-2pt, topsep=1pt, leftmargin=12pt]
\item \textbf{RQ1}: How can the task completion ability of Agentic Systems be evaluated across a wide range of domains? 
\item \textbf{RQ2}: How can the quality and human-alignment of $Judge$, tasked with evaluating these agentic systems, be improved?
\end{itemize}

By addressing these questions, we aim to establish a general evaluation framework that reduces human dependency while maintaining semantic rigor and cross-domain applicability.

\paragraph{Contributions}
The key contributions of this work are:

\begin{itemize} [itemsep=-2pt, topsep=1pt, leftmargin=12pt]
\item We present a domain-agnostic decomposition of the agent evaluation task, enabling step-wise reasoning over task completion without presupposing task-specific heuristics.
\item We release an open-source, modular, and extensible evaluation framework equipped with interface-consistent modules that can be independently substituted or enhanced. A concrete instantiation demonstrates its compatibility and generalization across multiple domains.
\item Empirical results show that our implementation consistently outperforms existing baselines, achieving stronger alignment with human evaluations (see \autoref{fig:v3-preformance}).
\end{itemize}

\paragraph{Proposed Approach}
Our framework is inspired by how expert human evaluators decompose complex tasks into logical subgoals and assess correctness at each step via evidence-based reasoning.

The $Judge$ agent architecture incorporates multiple LLM-based and agentic modules that collectively verify both intermediate reasoning steps and final outputs of the $Actor$ agent. For our experiments, we instantiate the $Actor$ using the Magentic-One agent, a general-purpose agent capable of multi-step task execution. Architectural and implementation details are provided in Section~3.

We evaluate our framework on two benchmark datasets, GAIA and BigCodeBench, which focus on text-based tasks. Evaluation quality is assessed by comparing the $Judge$’s verdicts against human annotations to measure alignment and reliability.

\begin{figure*}[t!]
  \centering
   \includegraphics[width=0.85\linewidth]{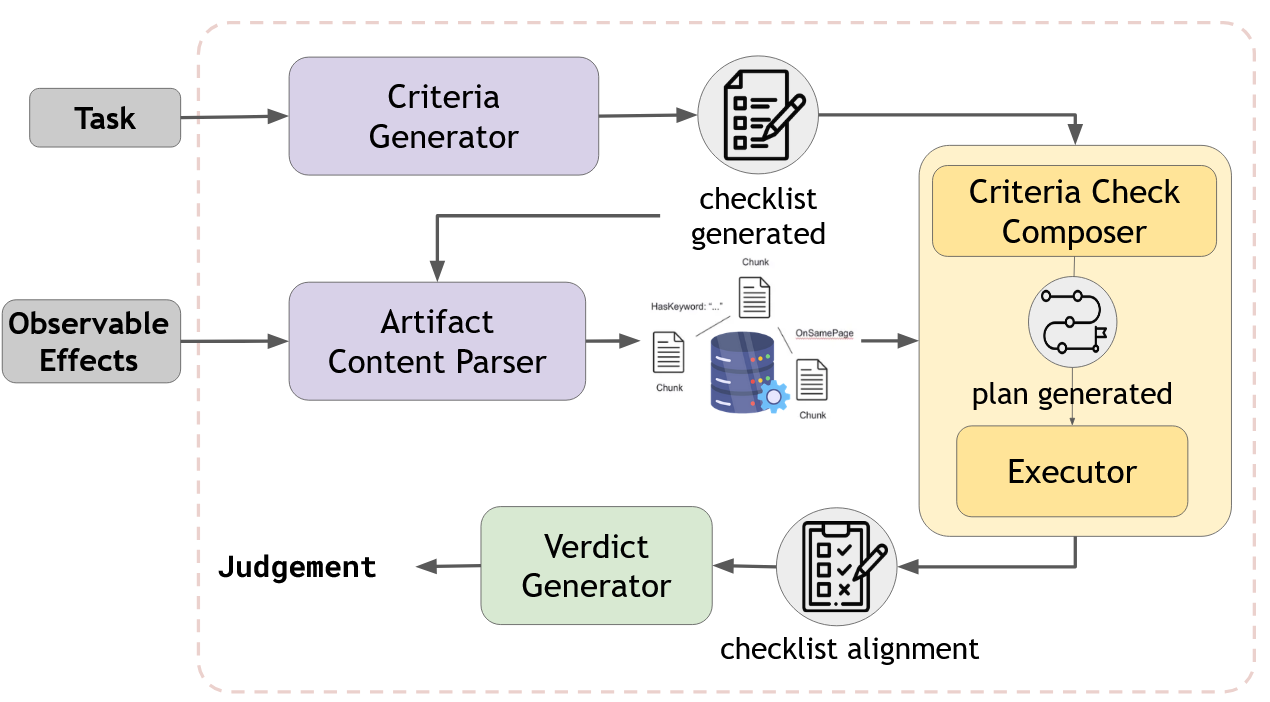}
   \caption{Proposed method overview. The input- $Task$ is comprised of not only query text, but also any type of attachment and tools that might be provided to the $Actor$.}
   \label{fig:revised-judge}
\end{figure*}

\section{Related Work}
Recent research has increasingly focused on evaluating large language models (LLMs) and autonomous agents by leveraging other LLMs or agents as evaluators, aiming to supplement or even replace human evaluation.

\paragraph{\textbf{LLM-as-a-Judge:}}
Several studies have shown that LLMs can serve as effective judges. For example, MT-Bench and Chatbot Arena \citep{zheng2023judgingllmasajudgemtbenchchatbot} use human-aligned LLMs to score open-ended responses, showing high agreement with human evaluations. Compass-Judger-1 \citep{cao2024compassjudger1allinonejudgemodel} trains an open-source judge using a mix of public and self-collected feedback data for scoring and pairwise comparisons. Prometheus \citep{kim2024prometheusinducingfinegrainedevaluation} further improves evaluation fidelity by training on a diverse feedback dataset, achieving results comparable to GPT-4.

While effective, these methods focus on static, text-based tasks. Our work builds on them by expanding evaluation to more complex, step-driven agentic behaviors.

\paragraph{\textbf{Multi-LLM as Judge Evaluation:}}
Multi-LLM setups simulate human-like debate to enhance evaluation quality. ChatEval \citep{chan2023chatevalbetterllmbasedevaluators} assigns roles to LLMs for collaborative judging, while AutoArena \citep{zhao2024autoarenaautomatingllmevaluations} uses peer debate and committee voting to select top responses. Though limited to chat tasks, these frameworks motivate our own $Judge$ design, aimed at evaluating broader workflows.

\paragraph{\textbf{Evaluating Agentic Systems:}}
Agent-as-a-Judge \citep{zhuge2024agentasajudgeevaluateagentsagents} introduces the idea of agent judging agents. However, it is constrained to DevAI dataset which contains ML tasks introduced through this paper. Moreover, this work needs a set of soft and hard requirements for each task as an input in order to perform evaluations, which hugely limits the extensibility of this method towards other datasets of same domain. Our work removes this requirement by automatically generating a set of checklist criteria agnostic to domain or dataset. 

\paragraph{\textbf{Step-by-Step Evaluation:}}
Since Agentic Systems often operate in sequences, evaluation should track not just final outputs but intermediate steps. LLM-Reasoners \citep{hao2024llmreasonersnewevaluation} proposes AutoRace to assess reasoning chains like Chain of Thought and Tree of Thought for LLMs. We adopt a similar philosophy, our $Judge$ agent evaluates the completeness and correctness of each step in the $Actor$ agent’s process, enabling more accurate and transparent assessments.

\section{Method}

\autoref{fig:revised-judge}. gives an overview of the proposed $Judge$ system. The remainder of this section describes the functional modules of the $Judge$ framework in detail.


\subsection{Criteria Generator}
Given a task description, the Criteria Generator module generates a concise list of checklist questions which must be met by the $Actor$ Agent to successfully complete the task. The generated checklist aims to verify whether the agent has satisfied the “explicit” requirements stated in the task, while minimizing assumptions. 

It follows detailed prompting guidelines to ensure that each checklist question is strictly aligned with the task. The questions are designed to be binary ("yes" or "no") in nature and focus on a single requirement at a time. This approach simplifies evaluation by avoiding complex, nested questions that would otherwise require additional reasoning or multiple steps.

\begin{figure}[ht!]
  \centering
   \includegraphics[width=1.0\linewidth]{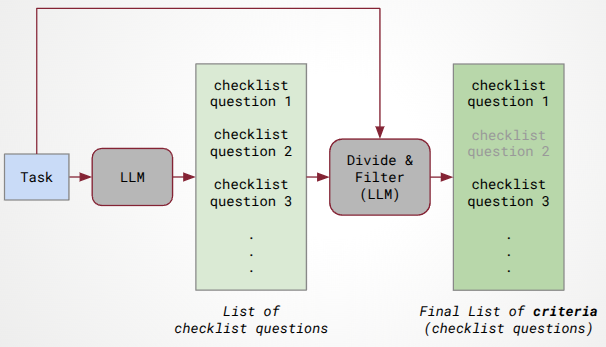}
   \caption{Architecture overview for Criteria Generator. The blue box represents input to the module and the green box depicts output.}
   \label{fig:cg}
\end{figure}

To further enhance relevance, an LLM-based filtering step is applied post-generation to remove questions that are either redundant or loosely connected to the task goal. \autoref{fig:cg} shows a comprehensive overview of the data flow for the implementation of this module.

\subsection{Artifact Content Parser}

This module is responsible for structuring and accurately retrieving relevant information ("proofs") from the log file generated by the $Actor$. 
The first step is handled by the \textbf{Indexer}, followed by the \textbf{Retriever}.

\begin{figure}[ht!]
  \centering
   \includegraphics[width=1.0\linewidth]{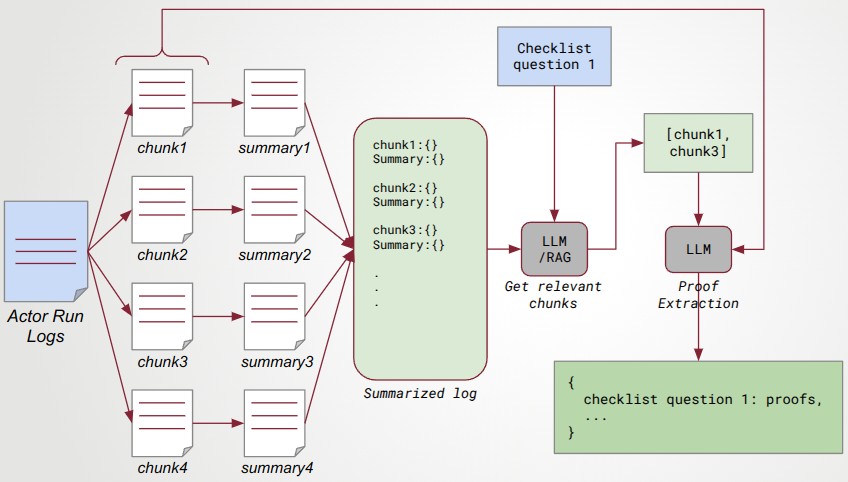}
   \caption{Architecture overview for Artifact Content Parser. The blue box represent input to the module and the green box depicts output.}
   \label{fig:ir}
\end{figure}

This entire pipeline is inspired by the Retrieval Augmented Generation (RAG) approach. \autoref{fig:ir} describes the overall flow.


\textbf{Indexer:}
It handles the efficient organization and structuring of log files (with current focus on a single log file) which is generally over 2000 lines.
It operates in two key steps:
\begin{enumerate}[itemsep=-2pt, topsep=1pt,leftmargin=12pt]
    \item \textbf{Chunking}: 
    The Indexer begins with chunking. There are two key motivations for this step. First, lengthy log files generated by the $Actor$ may exceed the context window of the language model. Second, even if the entire content of the log fits within the context window, chunking improves efficiency by reducing irrelevant information and optimizing latency and token usage. 
    

    We use 300-token chunks, as this configuration performs best in combination with a cross-encoder during retrieval.

    \item \textbf{Summarization}: Each chunk is summarized using an LLM to serve as a content index, which is important given the presence of mixed formats such as code, HTML tags, and plain text in the log. This improves performance significantly compared to using the raw chunks directly.
    
\end{enumerate}
The resulting summaries are passed to the next component, the \textbf{Retriever}.

The \textbf{Retriever} is responsible for identifying and extracting the relevant proofs from the log that align with specific checklist questions. This involves two steps:
\begin{enumerate}[itemsep=-2pt, topsep=1pt, leftmargin=12pt]
    \item \textbf{Chunk Retrieval}: Cross-encoder (ms-marco-MiniLM-L-6-v2) uses the checklist question against each generated summaries to identify the most relevant chunks. The relevant chunks here are chosen based on a set threshold.
    \item \textbf{Snippet Extraction}: For each relevant chunk, an LLM is employed to extract the precise snippet from the relevant chunks that serves as "proof" against each checklist question. These extracted snippets against each checklist question are passed to the subsequent module, the Criteria Check Composer.
\end{enumerate}



\subsection{Criteria Check Composer (C3)}

This module functions as the final integrative module responsible for synthesizing all upstream outputs into a coherent verification strategy. For each checklist query, C3 constructs an execution plan by referencing the relevant $Actor$ logs and using LLM-based reasoning alongside optionally using any externally provisioned computational assets such as structured knowledge bases, tools (like web search, file explorer, computer terminal, etc.).

These auxiliary resources may encompass annotator-derived procedural traces (e.g., intermediate steps recorded during dataset curation), canonical task solutions, predefined lookup tables, or any dataset-embedded metadata intended to support interpretability or correctness checking.

\begin{figure}[ht!]
  \centering
   \includegraphics[width=1.0\linewidth]{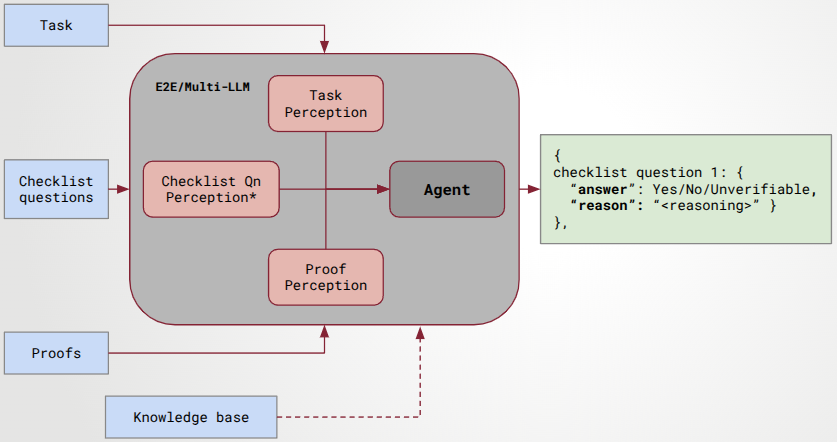}
   \caption{Architecture overview for Criteria Check Composer. The blue boxes represent inputs to the module (with dashed arrow as an optional input) and the green box depicts output. The dark gray box is the middle highlights the implicit/explicit components constituting this module}
   \label{fig:c3}
\end{figure}

To architect the verification pipeline, C3 leverages modular perceptual analyzers specific to each input modality or source. These submodules facilitate the decomposition, classification, and semantic parsing of input signals, enabling the system to instantiate a dynamic, decision tree based action graph. This graph encodes conditional verification pathways, allowing for fine-grained alignment between evidence sources and checklist criteria. \autoref{fig:c3} provides a schematic representation of the system's architecture.

\paragraph{\textit{Checklist Question Perception.}}
This component implements a hierarchical multi-stage classification pipeline over checklist-style queries, leveraging large language model (LLM) inference at each stage. At the first level, each question is classified into one of two primary semantic categories: $\mathbf{factual}$ or $\mathbf{logical}$. A query is labeled as $\mathbf{factual}$ if its resolution necessitates incorporation of external world knowledge or previously acquired domain understanding that is not inferable from the execution trace or log artifacts alone. Conversely, a $\mathbf{logical}$ query is one whose truth value or correctness can be determined strictly from the internal artifacts of task execution (e.g., function outputs, logs, or traces), potentially requiring auxiliary symbolic tools (e.g., code interpreters, calculators) or embedded logical reasoning mechanisms.

Subsequently, questions categorized as $\mathbf{logical}$ are further disambiguated into subtypes: $\mathbf{reasoning}$ or $\mathbf{coding}$. The $\mathbf{reasoning}$ class encapsulates tasks requiring abstract inference, deductive logic, or multi-step cognitive operations over the observed artifacts, while the $\mathbf{coding}$ class pertains to instances requiring programmatic analysis, code generation, or interpretation. Class labels are deliberately designed to be interpretable yet exhaustive within the operational scope of our task framework. Each classification stage is independently parameterized and executed through discrete LLM invocations to maintain modularity and compositionality.

\paragraph{\textit{Proof Perception.}} This sub-module conducts an automated verification pass to determine whether the candidate proofs associated with each checklist item are semantically sufficient for downstream evaluation. If the verification yields inconclusive or incomplete justification, the module triggers an expansion of the contextual retrieval window to incorporate broader evidence. This decision process is implemented via an LLM-based inference mechanism.

\begin{figure}[ht!]
  \centering
   \includegraphics[width=1.0\linewidth]{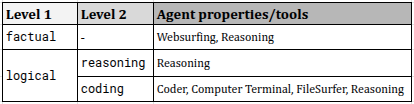}
   \caption{Checklist classes and handler of respective cases. The right most column signifies the bare minimum capabilities and/or expertise expected by the Agent handler}
   \label{fig:c3-decision}
\end{figure}

Conditioned on the inferred verification type, the system dispatches the corresponding evaluation task to either a specialized language model or a general-purpose LLM-driven agentic pipeline. As detailed in \autoref{fig:c3-decision}, verification tasks categorized as $\mathbf{factual}$ or $\mathbf{coding}$ are processed through a modular multi-agent architecture, Magentic-One. This framework comprises a planning agent (Orchestrator), a code generation agent, a retrieval/surfing agent, and a code execution environment. The planner synthesizes a verification trajectory which is subsequently executed against the retrieved artifacts. Conversely, $\mathbf{reasoning}$-type tasks are verified via a single-step LLM inference, as they predominantly involve abstract or logical inference chains that do not benefit from procedural decomposition.

\paragraph{\textit{Task Perception.}} 
Rather than constituting a discrete module, task perception functions as an overarching conceptual layer within the C3 pipeline. Its role is to inform verification planning by coupling checklist-based reasoning with a nuanced, semantic interpretation of the task description. To operationalize this, each LLM-driven sub-step is explicitly conditioned on the task under evaluation, embedding task-level awareness into the examination of both proofs and checklist items. This integration ensures that verification is contextually grounded and sensitive to the specificities of the task at hand.

\subsection{Verdict Generator}
\begin{figure}[ht!]
  \centering
   \includegraphics[width=1.0\linewidth]{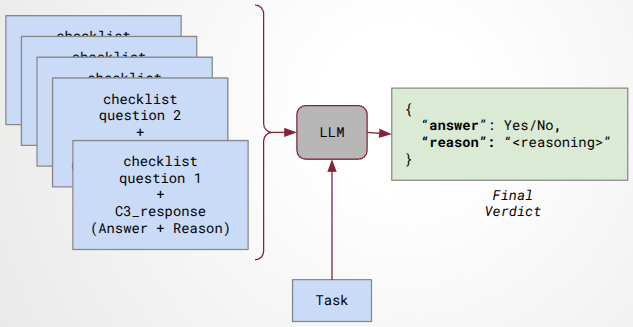}
   \caption{Architecture overview for Verdict Generator. The blue box represent input to the module and the green box depicts output.}
   \label{fig:vr}
\end{figure}

Finally, this module reasons with all the available information and outputs of previous modules to come to a verdict of whether the task was successfully completed by the $Actor$. For example: Given a task, list of checklist questions with their proofs that pertain to each checklist question, and the {output from each checklist from C3}, the verdict reasoner gives a single "Yes" or "No" as the answer. \autoref{fig:vr} shows the data flow for this module.

\section{Experimental Setup}
\subsection{Datasets}
To make a $Judge$ framework generalizable across multiple domains, it is important to include a variety of datasets representing diverse tasks. Since it is labor-intensive to annotate the $Actor$ results, we need to ensure that the datasets contain Auto-Evaluation techniques or Ground Truth to evaluate the $Actor$ agent. This will help us measure of how well our $Judge$ is able to evaluate or align with human evaluation.

\begin{table*}[ht]
\centering
\resizebox{0.95\textwidth}{!}{%
\begin{tabular}{|l|c|c|c|c|}
\hline
\multirow{2}{*}{\textbf{Metric}} & \multicolumn{2}{c|}{\textbf{GAIA (21 Pass, 21 Fail)}} & \multicolumn{2}{c|}{\textbf{Bigcode (28 Pass, 10 Fail)}} \\
\cline{2-5}
 & \textbf{LLM-as-a-Judge} & \textbf{Our-Judge (v3)} & \textbf{LLM-as-a-Judge} & \textbf{Our-Judge (v3)} \\
\hline
Accuracy & 57.14\% & \textbf{61.90\%}~$\uparrow$ & 63.16\% & \textbf{73.68\%}~$\uparrow$ \\
Precision & 60.87\% & \textbf{66.67\%}~$\uparrow$ & 76.47\% & \textbf{92.31\%}~$\uparrow$ \\
Recall & 60.87\% & \textbf{66.67\%}~$\uparrow$ & 100.00\% & 92.30\% \\
Specificity & 52.63\% & \textbf{57.14\%}~$\uparrow$ & 33.33\% & 33.33\% \\
\hline
\end{tabular}
}
\caption{Performance metrics of different judge models on the Inference Dataset}
\label{table:performance-metrics}
\end{table*}

\subsubsection{Data source}
\vspace{-2pt}
Our approach currently leverages \textbf{text-only} datasets such as GAIA \citep{mialon2023gaiabenchmarkgeneralai} and BigCodeBench \citep{zhuo2024bigcodebench}. GAIA provides general reasoning challenges involving multi-modality and web browsing tasks, while BigCodeBench evaluates LLMs on complex programming tasks requiring intermediate reasoning and function calls. 

All these datasets already exist and are publicly available through GitHub or HuggingFace.

\subsubsection{Data splits}
Since our focus is purely on evaluation, the dataset is sampled as follows:
\begin{itemize} [itemsep=-2pt, topsep=1pt, leftmargin=12pt]
    \item From GAIA, 42 out of 53 Level-1 tasks were selected, excluding those with multimodal attachments. A balanced distribution of $Actor$ runs was ensured across these tasks where 21 tasks were completed successfully and 21 were not. 
    
    \item {From BigCodeBench, 38 tasks were selected, excluding those with incorrect test-cases (which would have made the evaluation ambiguious). 28 tasks were completed successfully and 10 were not. }
\end{itemize}

\vspace{6pt}

\subsection{Evaluation Method}
For each dataset, three types of evaluations were considered:
\begin{enumerate} [itemsep=-2pt, topsep=1pt, leftmargin=12pt]
    \item \textbf{Human Evaluation:}
    \begin{itemize}  [itemsep=-2pt, topsep=1pt, leftmargin=12pt] 
        \item If ground truth was available in the dataset, it was used to guide the evaluation.
        \item If ground truth was not available, evaluators assessed correctness based on available evaluation methods, such as test cases.
        \item In the absence of both, annotators manually verified task completion through direct inspection.
    \end{itemize}
    \item \textbf{LLM as a Judge Verdict:}
    \begin{itemize} [itemsep=-2pt, topsep=1pt, leftmargin=12pt]
        \item A binary verdict (\texttt{Yes} or \texttt{No}) generated by GPT-4o-based LLM on the provided task description and final output.
    \end{itemize}
    \item \textbf{Judge Verdict:}
    \begin{itemize} [itemsep=-2pt, topsep=1pt, leftmargin=12pt]
        \item A binary verdict (\texttt{Yes} or \texttt{No}) produced by our framework, given the task description and the corresponding log file.
    \end{itemize}
\end{enumerate}

To measure evaluation quality, human alignment was calculated by constructing a confusion matrix between the human evaluation and the verdicts generated by each method. Human alignment with the LLM as a Judge serves as the baseline for evaluation. The human alignment score for the $Judge$ verdict is compared against this baseline to assess improvements.

\subsection{Baselines}
Due to the lack of prior research for a general agent as a judge across domains, there are no agentic baselines to consider, we implement LLM-as-a-Judge as the baseline. This choice stems from the hypothesis that agentic judge, which incorporates intermediate reasoning steps, will outperform a LLM-based judge that relies solely on final outputs.
To address the context window limitations of LLMs in LLM-as-a-Judge, we pass only the final output of the $Actor$ to the LLM-as-a-Judge, instead of verbose log files. By focusing solely on the final output, we effectively mitigate issues caused by the excessive length of detailed logs.

\section{Results and Discussions}
Our inital data setup lacked balanced $Actor$ runs due to the random nature of sampling (see \autoref{tab:old_data_setup} from Appendix). We address that in our current data setup by (1) increasing the sample size and (2) iteratively increasing the maximum number of turns allowed for the $Actor$ agent.
The results for the current run can be found in  \autoref{table:performance-metrics}. We can see that our system consistently beats the baseline, and performs better on the BigCodeBench dataset [\autoref{fig:v3-preformance}]. On BigCodeBench, the system seems to high precision \textbf{(92.31\%)}. From the other metrics, it can be concluded that the balanced dataset allowed for better visibility into the system's true capabilities.
\subsection{Error Analysis}

\begin{itemize} [itemsep=-2pt, topsep=1pt, leftmargin=12pt]
    \item Some evaluation tasks demand more than just a binary check. For example, given the task - 
    \begin{quote}
        "Of the authors (First M. Last) that worked on the paper 'sPie Menus or Linear Menus, Which Is Better?' in 2015, what was the title of the first paper authored by the one that had authored prior papers?"
    \end{quote}
    our existing judge framework bases its evaluations on evidence extracted from the actor agent’s execution logs. In this instance, the judge assigns positive verdict essentially trusting
    the validity of those logs, even though the actor’s answer does not match the ground truth. To preserve a clear distinction from the Actor Agent, our current framework prevents the judge from solving the tasks directly. However, the example highlights that it is imperative for a judge system to perform the task itself to validate answers concretely. 

    \item The Artifact Content Parser has mainly two failure modes. First, it occasionally injects opinions into its output; although undesirable, this behavior does not really impair the Criteria Check Composer (C3) module’s ability to verify claims. Second, the parser sometimes extracts proof snippets directly from the Actor Plan. In these instances, C3 interprets these proofs as established evidence of that ‘action’ being performed and consequently generates incorrect verdict.

    \item In some tasks that involve “role-playing” examples like: 
    \begin{quote}
        “You are Van Helsing, a renowned vampire hunter. A Count of Moldova, La»õcu IV, son of  Costea, has tasked you with investigating the village of »òirnea in neighboring Wallachia. The Count's advisors have reported that a vampire was spotted crossing the border near the village, and would like you to investigate it.”
    \end{quote}
    the Content Generator module often confuses the fictional role-play directives with the actor agent’s actual behavior. As a result, it may pose verification questions that assume in-world actions rather than monitoring what the agent genuinely executed. For example:
        \begin{itemize}
            \item Did the AI agent travel to the village of Șirnea as instructed?
            \item Did the AI agent observe the vampire jumping from rooftop to rooftop?
            \item Did the AI agent attempt to chase the vampire back to its home?
        \end{itemize}

Because these queries do not focus on the agent’s logged operations, the system cannot accurately verify whether the agent performed them.

\end{itemize}

\section{Conclusion}
In this work, we introduced a modular, domain-agnostic framework for evaluating agent task completion by assessing not just final outcomes, but the reasoning steps behind them. Our Judge Agent, evaluated on GAIA and BigCodeBench, achieved 4.76\% and 10.52\% higher alignment with human judgments compared to the GPT-4o LLM-as-a-Judge baseline.

These results highlight the strength of step-wise evaluation in capturing task completion in a more robust manner, offering a scalable foundation for evaluating increasingly capable agentic systems.

However, the current framework is not without limitations. It does not yet support the evaluation of multi-modal tasks, datasets, or domains. The Criteria Generator is currently limited to generating checklist questions for text-based tasks and does not handle tasks involving files or attachments. Likewise, the Artifact Content Parser can process only a single log file and does not account for multiple outputs or additional artifacts produced during task execution.

To address these gaps, we plan to extend the framework with an Environment Explorer module, integrated into the Criteria Check Composer, to support inspection of file-based outputs and more complex evaluation scenarios.

\section{Acknowledgments}
We would like to thank our mentors from University of Massachusetts Amherst and Microsoft AI Development Acceleration Program (MAIDAP). 

\bibliography{custom}

\onecolumn

\section{Appendix}
\subsection{Initial Data Setup and Results}

We initially used the following setup to iteratively improve our modules' implementation by error analysis and identifying bottlenecks.
\begin{itemize} [itemsep=-2pt, topsep=1pt, leftmargin=12pt]
    \item From GAIA, we randomly sampled 10 tasks for varying difficulty levels (1, 2, 3).
    \item From BigCodeBench, we randomly sampled 11 tasks.
\end{itemize}
\vspace{1em}

We measured the performance of our $Judge$ framework in terms of human alignment and overall classification performance. As shown in \autoref{fig:alignment-gaia}. several Judge configurations \textbf{(see \autoref{tab:module_version_details} in Appendix for more details on configurations)} outperform the baseline LLM-as-a-Judge in terms of human alignment.
\begin{figure}[ht]
    \centering
    \includegraphics[width=0.5\linewidth]{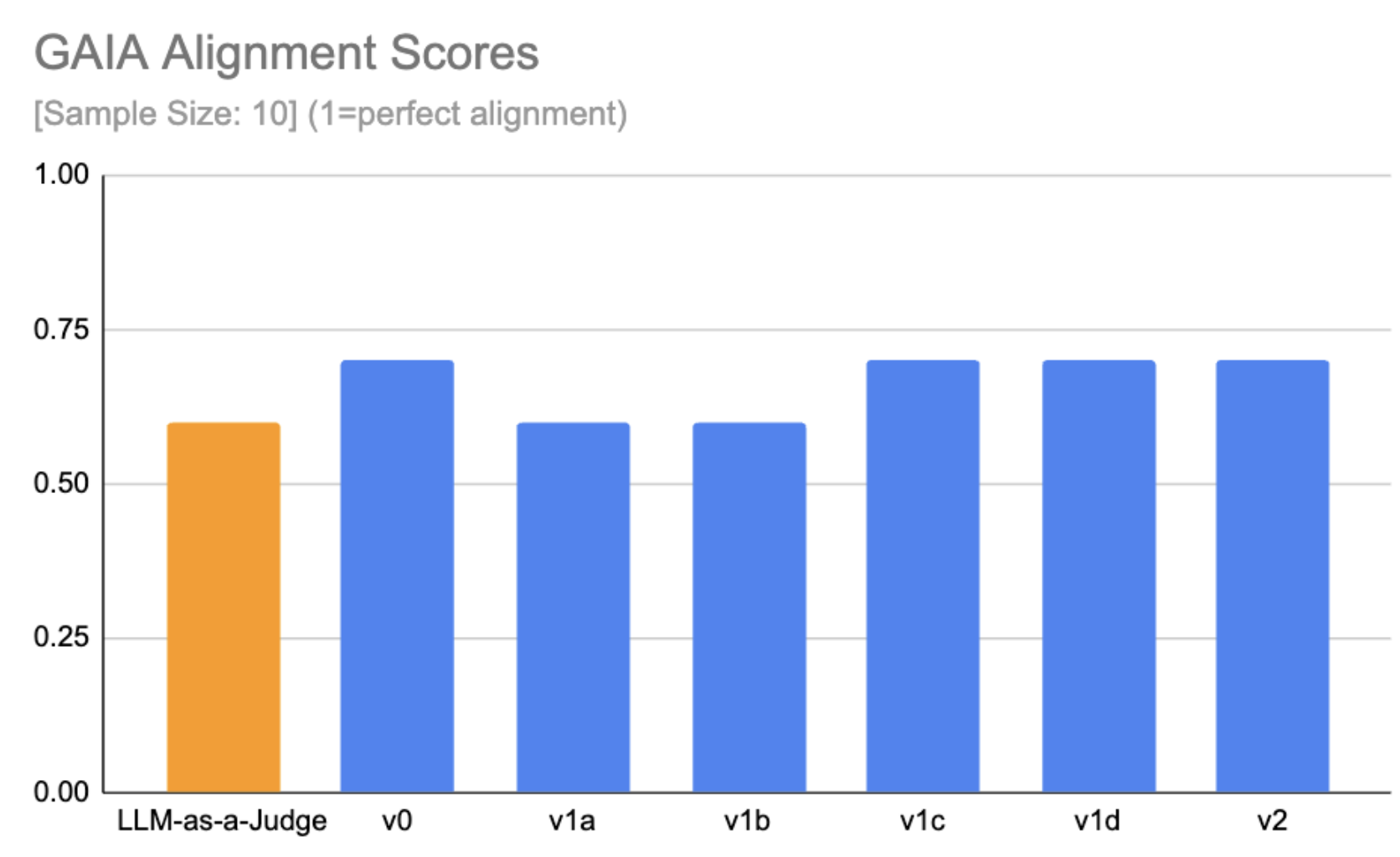}
    \caption{GAIA - Initial Data Setup - Alignment Scores}
    \label{fig:alignment-gaia}
\end{figure}
\FloatBarrier

\begin{figure} [ht]
    \centering
    \includegraphics[width=0.5\linewidth]{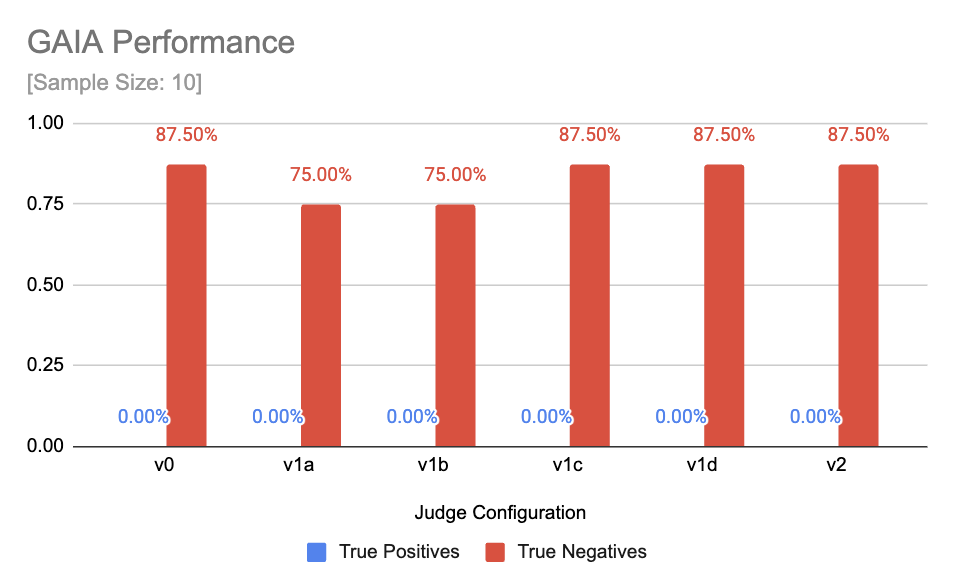}
    \caption{GAIA - Initial Data Setup - Classification Performance}
    \label{fig:classification-gaia}
\end{figure}

In \autoref{fig:classification-gaia}., we observed that none of the $Judge$ variants correctly classified any true positives. While this indicates that the $Judge$ agents are effective at rejecting incorrect completions, it also suggests an overly conservative behavior when it comes to crediting correct ones. One contributing factor is the imbalance in the GAIA samples used for Validation Setup. Only 2 out of the 10 tasks were successfully completed by the $Actor$ agent. This skew may have biased the $Judge$ toward negative classifications, highlighting the importance of using balanced evaluation sets for performance assessment.
\FloatBarrier

\begin{figure} [ht]
    \centering
    \includegraphics[width=0.5\linewidth]{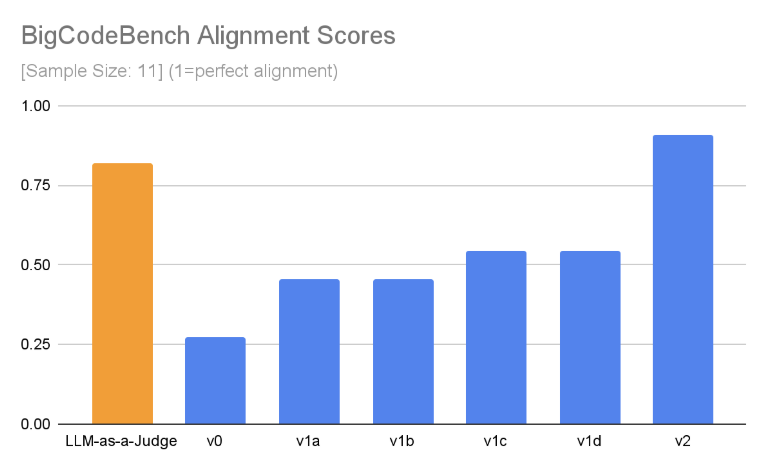}
    \caption{BigCodeBench- Initial Data Setup - Alignment Scores}
    \label{fig:alignment-bigcodebench}
\end{figure}

In \autoref{fig:alignment-bigcodebench}., interestingly the LLM-as-a-Judge variant outperforms most $Judge$ configurations. However, with each configuration of the $Judge$, the human alignment steadily improves and version v2 exceeds the baseline. 
\FloatBarrier

\begin{figure}[ht]
    \centering
    \includegraphics[width=0.5\linewidth]{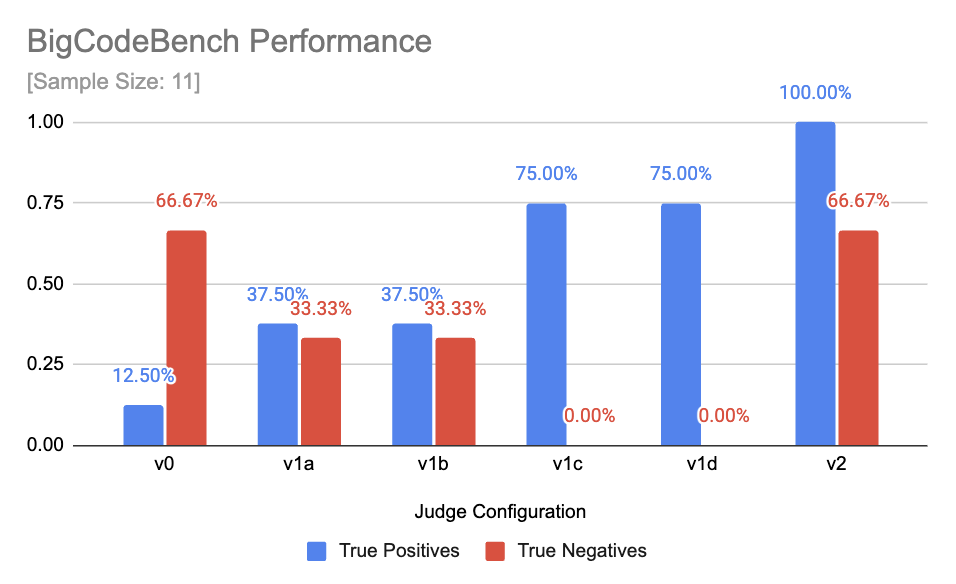}
    \caption{BigCodeBench - Initial Data Setup - Classification Performance}
    \label{fig:classification-bigcodebench}
\end{figure}

In \autoref{fig:classification-bigcodebench}., a similar increasing trend is observed with each configuration of the $Judge$ agent. Version v0 and v1a-v1b seem to perform poorly in identifying correct completions (true positives), whereas versions v1c and v1d improve significantly with 75\% true positive rates. Version v2 beats all the previous versions and achieves a perfect true positive rate of 100\%.
\FloatBarrier

\begin{figure}[ht]
    \centering
    \includegraphics[width=0.5\linewidth]{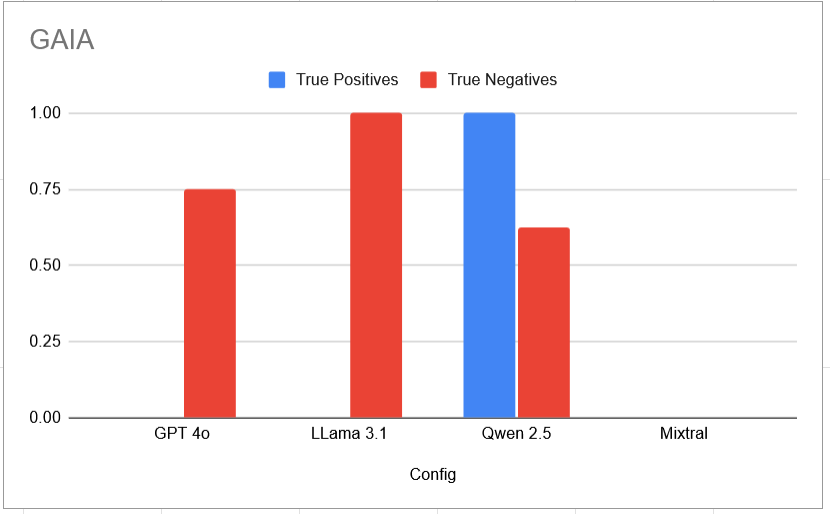}
    \caption{GAIA - Initial Data Setup - Open Source Models Comparison}
    \label{fig:oss_llm-gaia}
\end{figure}

\begin{figure}[ht]
    \centering
    \includegraphics[width=0.5\linewidth]{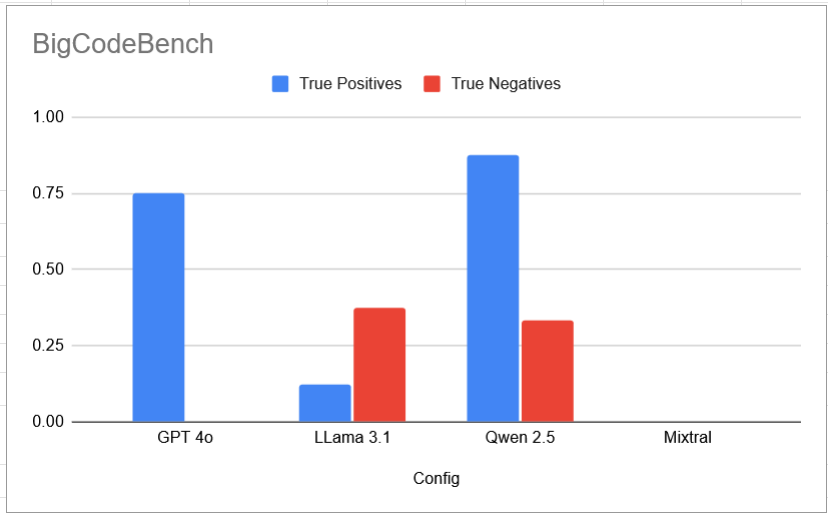}
    \caption{BigCodeBench - Initial Data Setup - Open Source Models Comparison}
    \label{fig:oss_llm-bigcodebench}
\end{figure}

In \autoref{fig:oss_llm-gaia}. and \autoref{fig:oss_llm-bigcodebench}., we compared the performance of our judge system against various underlying popular LLMs, many of which are Open Source. The primary intent was to determine the right model to power our judge system. While at first glance, it may seem that the Qwen 2.5 model performed better amongst all of the tested models, on deeper inspection of the judge logs, it was found that the Qwen model demonstrated significant weakness in effective reasoning, producing inferior checklist questions, as compared to Open AI's GPT 4o.
\FloatBarrier

\FloatBarrier

\begin{table}[ht]
\centering
\resizebox{\textwidth}{!}{%
\begin{tabular}{|l|cccc|cccc|}
\hline
\multirow{2}{*}{\textbf{Judge}} & \multicolumn{4}{c|}{\textbf{GAIA (Actor: 2 Pass, 8 Fail)}} & \multicolumn{4}{c|}{\textbf{Bigcode (Actor: 8 Pass, 3 Fail)}} \\
 & Accuracy & Precision & Recall & Specificity & Accuracy & Precision & Recall & Specificity \\
\hline
LLM-as-a-Judge & 60.00\% & 42.86\% & 100.00\% & 42.86\% & 81.81\% & 81.81\% & 100.00\% & 0.00\% \\
\hline
v0 & 70.00\%~$\uparrow$ & 0.00\% & 0.00\% & 87.50\%~$\uparrow$ & 27.27\% & 50.00\% & 12.50\% & 66.67\%~$\uparrow$ \\
LLama 3.2 v0 & 60.00\% & 25.00\% & 50.00\% & 62.50\%~$\uparrow$ & 54.55\% & 66.67\% & 75.00\% & 0.00\% \\
v1a & 60.00\% & 0.00\% & 0.00\% & 75.00\%~$\uparrow$ & 45.45\% & 75.00\% & 37.50\% & 66.67\%~$\uparrow$ \\
v1b & 60.00\% & 0.00\% & 0.00\% & 75.00\%~$\uparrow$ & 45.45\% & 75.00\% & 50.00\% & 33.33\% \\
v1c & 70.00\%~$\uparrow$ & 0.00\% & 0.00\% & 87.50\%~$\uparrow$ & 54.55\% & 66.67\% & 75.00\% & 0.00\% \\
v1d & 70.00\%~$\uparrow$ & 0.00\% & 0.00\% & 87.50\%~$\uparrow$ & 54.55\% & 66.67\% & 75.00\% & 0.00\% \\
llama 3.2 v1d & \textbf{80.00\%}~$\uparrow$ & 0.00\% & 0.00\% & \textbf{100.00\%}~$\uparrow$ & 36.36\% & \textbf{100.00\%}~$\uparrow$ & 12.50\% & \textbf{100.00\%}~$\uparrow$ \\
qwen v1d & \textbf{80.00\%}~$\uparrow$ & \textbf{50.00\%}~$\uparrow$ & \textbf{100.00\%}~$\uparrow$ & 75.00\%~$\uparrow$ & 72.73\% & 77.78\% & 87.50\% & 33.33\%~$\uparrow$ \\
v2 & 70.00\%~$\uparrow$ & 0.00\% & 0.00\% & 87.50\%~$\uparrow$ & 90.91\%~$\uparrow$ & 88.89\%~$\uparrow$ & \textbf{100.00\%}~$\uparrow$ & 66.67\% \\
\hline
\end{tabular}
}
\caption{Performance metrics of different judge models on the Validation Dataset}
\label{tab:old_data_setup}
\end{table}

\subsection{Judge Version Configuration Details}
\begin{table}[ht]
    \centering
    \begin{tabular}{|c|p{6cm}|p{7cm}|}
        \hline
        \textbf{Judge Version} & \textbf{Description} & \textbf{Concern} \\
        \hline
        \hline
        v0 & Initial Implementation & Checklist has too many sanity checks\\
        \hline
        v1a & Upgraded Criteria Generator & Insufficient / Invalid Proof Extraction\\
        \hline
        v1b & Upgraded Artifact Content Retriever & Verification Process Failures\\
        \hline
        v1c & Upgraded Criteria Check Composer & Verdict Generation Lacking Context\\
        \hline
        v1d & Upgraded Verdict Reasoner & False Positives due to Improper Checklist\\
        \hline
        v2 & Upgraded Criteria Generator & Proofs Contain Values from Irrelevant Sections of the Actor Log\\
        \hline
        v3 & Upgraded Indexer Retriever & \\
        \hline
    \end{tabular}
    \caption{Judge Version Details}
    \label{tab:module_version_details}
\end{table}

\newpage

\subsection{GAIA example}
\subsubsection{Task}
\begin{lstlisting}
Where were the Vietnamese specimens described by Kuznetzov in Nedoshivina's 2010 paper eventually deposited? Just give me the city name without abbreviations.
\end{lstlisting}

\subsubsection{Judge Evaluation}
\begin{lstlisting}
{
    "verdict": "no",
    "eval": [
        {
            "question": "Did the agent provide a city name as the answer?",
            "proofs": {
                "proofs": "``` \nfile_0.txt = ---------- MagenticOneOrchestrator ----------\n\nTo answer this request we have assembled the following team:\n\n...\n- Once the paper is found, extract the relevant sections to identify the city in which the specimens were deposited. \n- Cross-check any findings to ensure the information is accurate and directly sourced from the paper.\n```\n\n``` \nfile_6.txt = Favorite\nShare\nFlag\ntexts\nA catalogue of type specimens of the Tortricidae described by V. I. Kuznetzov from Vietnam and deposited in the Zoological Institute, St. Petersburg\n```",
                "final_answer": "St. Petersburg"
            },
            "c3_response": {
                "answer": "yes",
                "reason": "The relevant artifact 'file_6.txt' specifies that the specimens described by Kuznetzov from Vietnam were deposited in the Zoological Institute, which is located in St. Petersburg. Therefore, the city name 'St. Petersburg' was provided as the answer.",
                "decision_path": [
                    "proofs+final_answer",
                    "logical",
                    "reasoning"
                ]
            }
        },
        {
            "question": "Was the city name given without any abbreviations?",
            "proofs": {
                "proofs": "```plaintext\n---------- MagenticOneOrchestrator ----------\nThe Vietnamese specimens described by Kuznetzov in Nedoshivina's 2010 paper were eventually deposited in St. Petersburg.\n```",
                "final_answer": "St. Petersburg"
            },
            "c3_response": {
                "answer": "no",
                "reason": "The city name provided in the relevant artifact is 'St. Petersburg', which is an abbreviated form of 'Saint Petersburg', thus not given without abbreviations.",
                "decision_path": [
                    "proofs",
                    "logical",
                    "reasoning"
                ]
            }
        },
        {
            "question": "Did the agent identify the city where the Vietnamese specimens described by Kuznetzov in Nedoshivina's 2010 paper were deposited?",
            "proofs": {
                "proofs": "```\nA huge amount of lepidopterous material was collected by KuzNETzov in Vietnam and all of it is now deposited in the collection of ZISP. \n```\n",
                "final_answer": "St. Petersburg"
            },
            "c3_response": {
                "answer": "yes",
                "reason": "The relevant artifact states that the material collected by Kuznetzov in Vietnam is now deposited in the collection of ZISP. Since ZISP is located in St. Petersburg, the answer to the checklist question is yes, as the city where the specimens were deposited is identified as St. Petersburg.",
                "decision_path": [
                    "proofs+final_answer",
                    "logical",
                    "reasoning"
                ]
            }
        }
    ]
}  
\end{lstlisting}

\newpage

\subsection{BigCode example}
\subsubsection{Task}
\begin{lstlisting}[language=Python]
from sklearn.preprocessing import StandardScaler
import seaborn as sns
import matplotlib.pyplot as plt


def task_func(df):
    """
    Standardize numeric columns in a DataFrame and return the heatmap of the correlation matrix. Missing values are replaced by the column's average.

    Parameters:
    - df (pandas.DataFrame): The pandas DataFrame to be standardized.

    Returns:
    - DataFrame: The pandas DataFrame after standardization.
    - Axes: A heatmap of the correlation matrix.

    Requirements:
    - sklearn.preprocessing.StandardScaler
    - seaborn
    - matplotlib.pyplot

    Example:
    >>> import pandas as pd
    >>> import numpy as np
    >>> df = pd.DataFrame([[1,2,3],[4,5,6],[7.0,np.nan,9.0]], columns=["c1","c2","c3"])
    >>> standardized_df, heatmap = task_func(df)
    >>> print(standardized_df)
             c1        c2        c3
    0 -1.224745 -1.224745 -1.224745
    1  0.000000  1.224745  0.000000
    2  1.224745  0.000000  1.224745
    """
\end{lstlisting}

\subsubsection{Judge Evaluation}
\begin{lstlisting}
{
    "verdict": "yes",
    "eval": [
        {
            "question": "Was the StandardScaler from sklearn.preprocessing used to standardize the numeric columns of the DataFrame?",
            "proofs": {
                "proofs": "```python\nimport pandas as pd\nfrom sklearn.preprocessing import StandardScaler\nimport seaborn as sns\nimport matplotlib.pyplot as plt\n\ndef task_func(df):\n    # Replace missing values with the column's average\n    df_filled = df.apply(lambda col: col.fillna(col.mean()), axis=0)\n    \n    # Standardize numeric columns\n    scaler = StandardScaler()\n    df_standardized = pd.DataFrame(scaler.fit_transform(df_filled), columns=df_filled.columns)\n    \n    # Generate the correlation matrix\n    corr_matrix = df_standardized.corr()\n    \n    # Plot the heatmap\n    plt.figure(figsize=(8, 6))\n    heat_map = sns.heatmap(corr_matrix, annot=True, cmap='coolwarm', fmt=\".2f\")\n    plt.title('Correlation Matrix Heatmap')\n    plt.show()\n    \n    return df_standardized, heat_map\n\n# Example usage\nif __name__ == \"__main__\":\n    import numpy as np\n    \n    df = pd.DataFrame([[1, 2, 3], [4, 5, 6], [7.0, np.nan, 9.0]], columns=[\"c1\", \"c2\", \"c3\"])\n    standardized_df, heatmap = task_func(df)\n    print(standardized_df)\n```",
                "final_answer": "import pandas as pd\nimport numpy as np\nfrom sklearn.preprocessing import StandardScaler\nimport seaborn as sns\nimport matplotlib.pyplot as plt\n\ndef task_func(df):\n    # Replace missing values with the column's average\n    df_filled = df.apply(lambda col: col.fillna(col.mean()), axis=0)\n    \n    # Standardize numeric columns\n    scaler = StandardScaler()\n    df_standardized = pd.DataFrame(scaler.fit_transform(df_filled), columns=df_filled.columns)\n    \n    # Generate the correlation matrix\n    corr_matrix = df_standardized.corr()\n    \n    # Plot the heatmap\n    plt.figure(figsize=(8, 6))\n    heat_map = sns.heatmap(corr_matrix, annot=True, cmap='coolwarm', fmt=\".2f\")\n    plt.title('Correlation Matrix Heatmap')\n    plt.show()\n    \n    return df_standardized, heat_map\n\n# Example usage\nif __name__ == \"__main__\":\n    df = pd.DataFrame([[1, 2, 3], [4, 5, 6], [7.0, np.nan, 9.0]], columns=[\"c1\", \"c2\", \"c3\"])\n    standardized_df, heatmap = task_func(df)\n    print(standardized_df)"
            },
            "c3_response": {
                "answer": "yes",
                "reason": "The relevant artifact shows that the StandardScaler from sklearn.preprocessing is instantiated and used to fit and transform the numeric columns of the DataFrame. This standardization process normalizes the columns to have a mean of 0 and a standard deviation of 1, as required by the checklist question.",
                "decision_path": [
                    "proofs",
                    "logical",
                    "reasoning"
                ]
            }
        },
        {
            "question": "Were missing values in the DataFrame replaced with the column's average?",
            "proofs": {
                "proofs": "```python\n# Replace missing values with the column's average\ndf_filled = df.apply(lambda col: col.fillna(col.mean()), axis=0)\n```",
                "final_answer": "import pandas as pd\nimport numpy as np\nfrom sklearn.preprocessing import StandardScaler\nimport seaborn as sns\nimport matplotlib.pyplot as plt\n\ndef task_func(df):\n    # Replace missing values with the column's average\n    df_filled = df.apply(lambda col: col.fillna(col.mean()), axis=0)\n    \n    # Standardize numeric columns\n    scaler = StandardScaler()\n    df_standardized = pd.DataFrame(scaler.fit_transform(df_filled), columns=df_filled.columns)\n    \n    # Generate the correlation matrix\n    corr_matrix = df_standardized.corr()\n    \n    # Plot the heatmap\n    plt.figure(figsize=(8, 6))\n    heat_map = sns.heatmap(corr_matrix, annot=True, cmap='coolwarm', fmt=\".2f\")\n    plt.title('Correlation Matrix Heatmap')\n    plt.show()\n    \n    return df_standardized, heat_map\n\n# Example usage\nif __name__ == \"__main__\":\n    df = pd.DataFrame([[1, 2, 3], [4, 5, 6], [7.0, np.nan, 9.0]], columns=[\"c1\", \"c2\", \"c3\"])\n    standardized_df, heatmap = task_func(df)\n    print(standardized_df)"
            },
            "c3_response": {
                "answer": "yes",
                "reason": "The relevant artifact shows that missing values in the DataFrame are replaced using the lambda function `col.fillna(col.mean())`, which fills in missing values with the column's average (mean). This aligns with the checklist requirement.",
                "decision_path": [
                    "proofs",
                    "logical",
                    "reasoning"
                ]
            }
        },
        {
            "question": "Was a heatmap of the correlation matrix generated using seaborn?",
            "proofs": {
                "proofs": "```python\nimport pandas as pd\nfrom sklearn.preprocessing import StandardScaler\nimport seaborn as sns\nimport matplotlib.pyplot as plt\n\ndef task_func(df):\n    # Replace missing values with the column's average\n    df_filled = df.apply(lambda col: col.fillna(col.mean()), axis=0)\n    \n    # Standardize numeric columns\n    scaler = StandardScaler()\n    df_standardized = pd.DataFrame(scaler.fit_transform(df_filled), columns=df_filled.columns)\n    \n    # Generate the correlation matrix\n    corr_matrix = df_standardized.corr()\n    \n    # Plot the heatmap\n    plt.figure(figsize=(8, 6))\n    heat_map = sns.heatmap(corr_matrix, annot=True, cmap='coolwarm', fmt=\".2f\")\n    plt.title('Correlation Matrix Heatmap')\n    plt.show()\n    \n    return df_standardized, heat_map\n\n# Example usage\nif __name__ == \"__main__\":\n    import numpy as np\n    \n    df = pd.DataFrame([[1, 2, 3], [4, 5, 6], [7.0, np.nan, 9.0]], columns=[\"c1\", \"c2\", \"c3\"])\n    standardized_df, heatmap = task_func(df)\n    print(standardized_df)\n```",
                "final_answer": "import pandas as pd\nimport numpy as np\nfrom sklearn.preprocessing import StandardScaler\nimport seaborn as sns\nimport matplotlib.pyplot as plt\n\ndef task_func(df):\n    # Replace missing values with the column's average\n    df_filled = df.apply(lambda col: col.fillna(col.mean()), axis=0)\n    \n    # Standardize numeric columns\n    scaler = StandardScaler()\n    df_standardized = pd.DataFrame(scaler.fit_transform(df_filled), columns=df_filled.columns)\n    \n    # Generate the correlation matrix\n    corr_matrix = df_standardized.corr()\n    \n    # Plot the heatmap\n    plt.figure(figsize=(8, 6))\n    heat_map = sns.heatmap(corr_matrix, annot=True, cmap='coolwarm', fmt=\".2f\")\n    plt.title('Correlation Matrix Heatmap')\n    plt.show()\n    \n    return df_standardized, heat_map\n\n# Example usage\nif __name__ == \"__main__\":\n    df = pd.DataFrame([[1, 2, 3], [4, 5, 6], [7.0, np.nan, 9.0]], columns=[\"c1\", \"c2\", \"c3\"])\n    standardized_df, heatmap = task_func(df)\n    print(standardized_df)"
            },
            "c3_response": {
                "answer": "yes",
                "reason": "The relevant artifact shows that a heatmap of the correlation matrix is generated using seaborn. The heatmap is created with sns.heatmap() using the correlation matrix, and the plt.show() function is called to display it.",
                "decision_path": [
                    "proofs",
                    "logical",
                    "reasoning"
                ]
            }
        },
        {
            "question": "Was matplotlib.pyplot used to display the heatmap?",
            "proofs": {
                "proofs": "```python\n# Plot the heatmap\nplt.figure(figsize=(8, 6))\nheat_map = sns.heatmap(corr_matrix, annot=True, cmap='coolwarm', fmt=\".2f\")\nplt.title('Correlation Matrix Heatmap')\nplt.show()\n```",
                "final_answer": "import pandas as pd\nimport numpy as np\nfrom sklearn.preprocessing import StandardScaler\nimport seaborn as sns\nimport matplotlib.pyplot as plt\n\ndef task_func(df):\n    # Replace missing values with the column's average\n    df_filled = df.apply(lambda col: col.fillna(col.mean()), axis=0)\n    \n    # Standardize numeric columns\n    scaler = StandardScaler()\n    df_standardized = pd.DataFrame(scaler.fit_transform(df_filled), columns=df_filled.columns)\n    \n    # Generate the correlation matrix\n    corr_matrix = df_standardized.corr()\n    \n    # Plot the heatmap\n    plt.figure(figsize=(8, 6))\n    heat_map = sns.heatmap(corr_matrix, annot=True, cmap='coolwarm', fmt=\".2f\")\n    plt.title('Correlation Matrix Heatmap')\n    plt.show()\n    \n    return df_standardized, heat_map\n\n# Example usage\nif __name__ == \"__main__\":\n    df = pd.DataFrame([[1, 2, 3], [4, 5, 6], [7.0, np.nan, 9.0]], columns=[\"c1\", \"c2\", \"c3\"])\n    standardized_df, heatmap = task_func(df)\n    print(standardized_df)"
            },
            "c3_response": {
                "answer": "yes",
                "reason": "matplotlib.pyplot was used to display the heatmap as indicated by the call to plt.show() following the seaborn heatmap creation, which utilizes matplotlib's plotting capabilities.",
                "decision_path": [
                    "proofs",
                    "logical",
                    "reasoning"
                ]
            }
        },
        {
            "question": "Did the function return both the standardized DataFrame and the Axes of the heatmap?",
            "proofs": {
                "proofs": "```python\ndef task_func(df):\n    # Replace missing values with the column's average\n    df_filled = df.apply(lambda col: col.fillna(col.mean()), axis=0)\n    \n    # Standardize numeric columns\n    scaler = StandardScaler()\n    df_standardized = pd.DataFrame(scaler.fit_transform(df_filled), columns=df_filled.columns)\n    \n    # Generate the correlation matrix\n    corr_matrix = df_standardized.corr()\n    \n    # Plot the heatmap\n    plt.figure(figsize=(8, 6))\n    heat_map = sns.heatmap(corr_matrix, annot=True, cmap='coolwarm', fmt=\".2f\")\n    plt.title('Correlation Matrix Heatmap')\n    plt.show()\n    \n    return df_standardized, heat_map\n\n# Example usage\nif __name__ == \"__main__\":\n    import numpy as np\n    \n    df = pd.DataFrame([[1, 2, 3], [4, 5, 6], [7.0, np.nan, 9.0]], columns=[\"c1\", \"c2\", \"c3\"])\n    standardized_df, heatmap = task_func(df)\n    print(standardized_df)\n```",
                "final_answer": "import pandas as pd\nimport numpy as np\nfrom sklearn.preprocessing import StandardScaler\nimport seaborn as sns\nimport matplotlib.pyplot as plt\n\ndef task_func(df):\n    # Replace missing values with the column's average\n    df_filled = df.apply(lambda col: col.fillna(col.mean()), axis=0)\n    \n    # Standardize numeric columns\n    scaler = StandardScaler()\n    df_standardized = pd.DataFrame(scaler.fit_transform(df_filled), columns=df_filled.columns)\n    \n    # Generate the correlation matrix\n    corr_matrix = df_standardized.corr()\n    \n    # Plot the heatmap\n    plt.figure(figsize=(8, 6))\n    heat_map = sns.heatmap(corr_matrix, annot=True, cmap='coolwarm', fmt=\".2f\")\n    plt.title('Correlation Matrix Heatmap')\n    plt.show()\n    \n    return df_standardized, heat_map\n\n# Example usage\nif __name__ == \"__main__\":\n    df = pd.DataFrame([[1, 2, 3], [4, 5, 6], [7.0, np.nan, 9.0]], columns=[\"c1\", \"c2\", \"c3\"])\n    standardized_df, heatmap = task_func(df)\n    print(standardized_df)"
            },
            "c3_response": {
                "answer": "yes",
                "reason": "The function 'task_func' returns both the standardized DataFrame and the Axes of the heatmap as indicated by 'return df_standardized, heat_map'. The standardized DataFrame is created using the StandardScaler from sklearn after filling missing values, and the heatmap is generated using seaborn and is returned as 'heat_map', which are the correct outputs according to the checklist question's requirements.",
                "decision_path": [
                    "proofs",
                    "logical",
                    "reasoning"
                ]
            }
        }
    ]
}
\end{lstlisting}

\end{document}